\documentclass{article}

\usepackage[dblblindworkshop,final]{neurips_2025}
\usepackage{graphicx}
 \workshoptitle{Evaluating the Evolving LLM Lifecycle: Benchmarks, Emergent Abilities, and Scaling}
\usepackage[utf8]{inputenc} 
\usepackage[T1]{fontenc}    
\usepackage{hyperref}       
\usepackage{url}            
\usepackage{booktabs}       
\usepackage{amsfonts}       
\usepackage{nicefrac}       
\usepackage{microtype}      
\usepackage{xcolor}         
\usepackage{float}
\usepackage[final]{changes}

\title{Medical AI Consensus: A Multi-Agent Framework for Radiology Report Generation and Evaluation}

\author{%
  Ahmed T. Elboardy\thanks{Corresponding Author} \\
  Graduate School of Information Science,\\
  University of Hyogo,\\
  Kobe 650-0047, Japan \\
  \texttt{af25s001@guh.u-hyogo.ac.jp } \\
   \And
   Ghada Khoriba \\
   Center for Informatics Science, \\
   School of Information Technology and Computer Science, \\
   Nile University,\\
   Giza 12588, Egypt \\
   \texttt{ghadakhoriba@nu.edu.eg} \\
   \AND
   Essam A. Rashed \\
   Graduate School of Information Science,\\
  University of Hyogo,\\
  Kobe 650-0047, Japan \\
  \texttt{rashed@gsis.u-hyogo.ac.jp } \\
}

\begin{document}
\maketitle
\begin{abstract}

Automating radiology report generation poses a dual challenge: building clinically reliable systems and designing rigorous evaluation protocols. We introduce a multi-agent reinforcement learning framework that serves as both a benchmark and evaluation environment for multimodal clinical reasoning in the radiology ecosystem. The proposed framework integrates large language models (LLMs) and large vision models (LVMs) within a modular architecture composed of ten specialized agents responsible for image analysis, feature extraction, report generation, review, and evaluation. This design enables fine-grained assessment at both the agent level (e.g., detection and segmentation accuracy) and the consensus level (e.g., report quality and clinical relevance). We demonstrate an implementation using chatGPT-4o on public radiology datasets, where LLMs act as evaluators alongside medical radiologist feedback. By aligning evaluation protocols with the LLM development lifecycle, including pretraining, finetuning, alignment, and deployment, the proposed benchmark establishes a path toward trustworthy deviance-based radiology report generation.
\end{abstract}

\section{Introduction and Motivation}

The integration of Large Language Models (LLMs) and Large Vision Models (LVMs) presents significant opportunities for advancing real-time medical report generation and evaluation~\cite{busch2025large, lu2024multimodal, tanno2025collaboration}. However, their deployment at scale requires structured coordination, which can be effectively achieved through robust Multi-Agent System (MAS) architectures~\cite{ferber2025development}. While qualitative discussions of MAS in medical imaging are available~\cite{feng2025m}, systematic quantitative benchmarks remain scarce, limiting rigorous evaluation across the LLM lifecycle~\cite{bedi2025testing}. To address this gap, we propose a multi-agent benchmarking framework designed to evaluate LLMs and LVMs throughout an end-to-end pipeline of radiology report generation. The proposed approach introduces a generative AI agentic architecture in which specialized agents collaborate under a central orchestrator. Each agent is assessed through task-specific metrics, while overall framework performance is evaluated via composite measures of report accuracy, clinical completeness, and human-in-the-loop review.

As a case study, we curated a benchmark dataset of multisequence brain MRI scans from cancer patients, annotated under the supervision of board-certified radiologists. This dataset supports both the training and validation of agentic evaluation strategies. By systematically quantifying performance at both the agent and framework levels, our benchmark advances the evolving LLM lifecycle-spanning fine-tuning, alignment, and clinical deployment. The proposed framework not only streamlines radiological workflows but also strengthens trust in generative AI systems by enabling transparent, reproducible, and clinically relevant evaluation of LLM-LVM integration.

\section{Proposed Framework}
\begin{figure}
    \centering
    \includegraphics[width=0.8\textwidth]{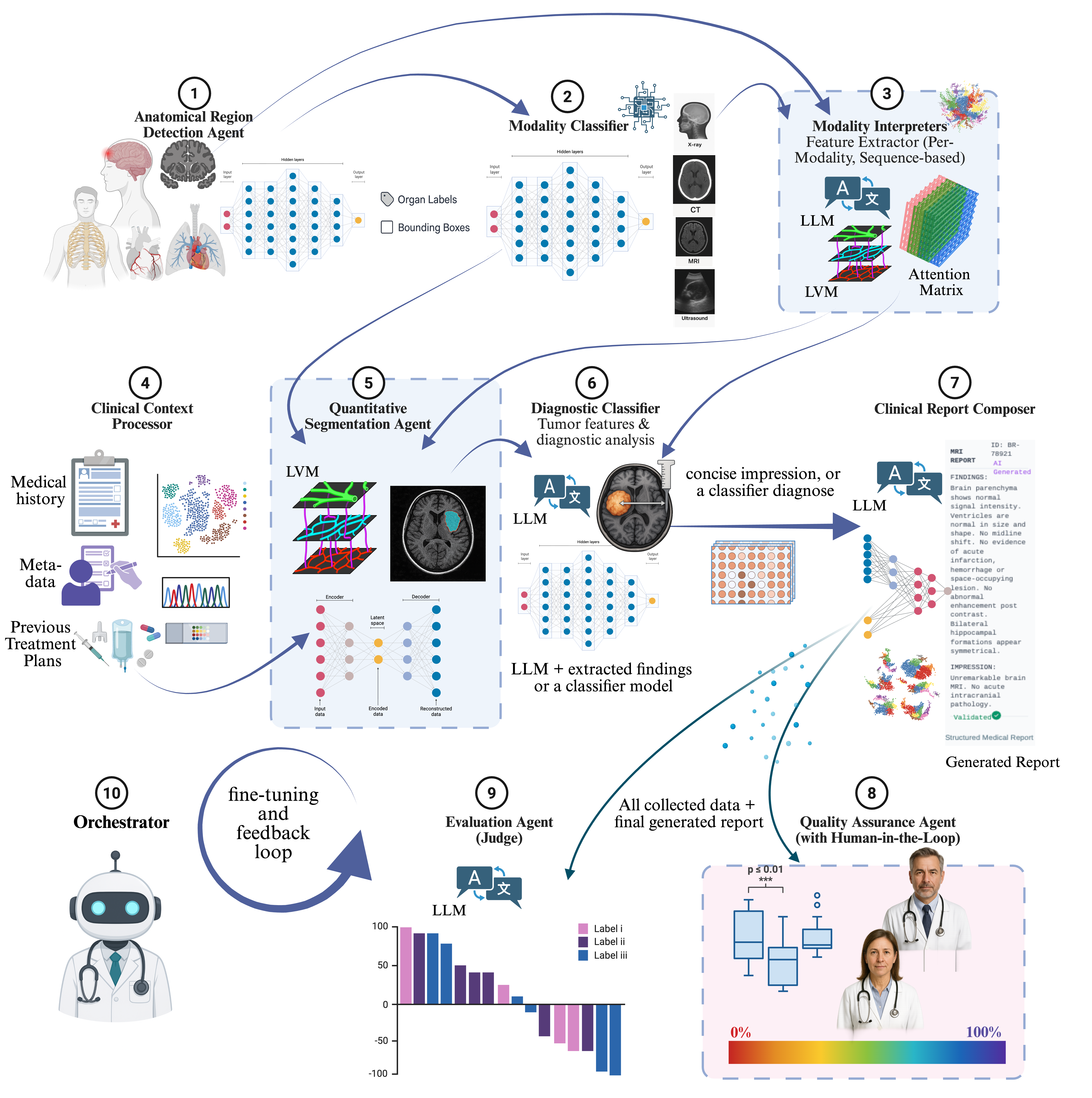}
    \caption{The proposed Medical AI Consensus: A Multi-Agent Framework. Ten specialized agents collaborate to generate radiology reports: (1) Anatomical Region Detection Agent, (2) Modality Classifier (determines imaging type), (3) Modality Interpreters (extract findings per modality), (4) Clinical Context Processor (analyzes patient metadata), (5) Quantitative Segmentation (measures/segment abnormalities), (6) Diagnostic Classifier (provides diagnostic analysis), (7) Clinical Report Composer (generate structured reports), (8) Quality Assurance Agent, (9) Evaluation Judge (automated LLM scoring), and (10) Orchestrator (manages workflow and validation).}
    \label{fig:placeholder}
\end{figure}

Figure~\ref{fig:placeholder} illustrates the architecture of the proposed multi-agent framework.  The pipeline is composed of ten specialized agents, each dedicated to a specific stage in real-time radiology image interpretation and report generation. These agents interacts in an iterative and cooperative manner, coordinated by a central orchestrator agent that ensures consistency and efficiency across the workflow. The framework is designed to be model-agnostic, allowing researchers to seamlessly integrate novel models and systematically evaluate their contributions within a standardized environment designed to simulate the real-life radiology ecosystem. The following provide a detailed description of the core agents and their respective roles within the framework.

{\bf Anatomical Region Detection Agent:}
This agent processes the input medical images, potentially spanning multiple sequences or modalities, to identify the anatomical region(s) and their spatial orientation. Its performance can be quantitatively assessed by comparing predicted regions against established anatomical labels.

{\bf Modality Classifier:} The modality classifier determines the imaging modality (e.g., X-ray, CT, MRI, US) and, when applicable, its specific subtype. Accurate modality recognition is essential, as reporting conventions and clinically relevant findings vary significantly across modalities. The evaluation of this component is conducted using standard classification accuracy metrics.

{\bf Modality Interpreters:} This category encompasses a pool of specialized agents, each tailored to a specific organ-modality combination (e.g., Chest X-ray Interpreter, Brain MRI Interpreter). Each agent, implemented as either an LLM or VLM, is optimized to extract modality-specific clinical features, including salient observations, abnormalities, quantitative measurements, and descriptive attributes. Performance is evaluated using task-specific metrics aligned with the clinical relevance and expected findings.

{\bf Clinical Context Processor:} Radiological studies are typically accompanied by rich metadata, including patient demographics, treatment history, clinical indications for the examination, earlier findings. The Clinical Context Processor is responsible for parsing, interpreting, and synthesizing the information to provide contextual input for the report generation process. Its performance can be evaluated by assessing the precision of structured field extraction, as well as, the accuracy and fidelity summarizing prior reports.

{\bf Quantitative Segmentation Agent:}
Upon detection of a clinically relevant abnormality by the modality interpreter agents, the orchestrator may invoke the quantitative segmentation agent to perform detailed delineation and quantification. Implemented as an LVM, this agent generates segmentation masks, bounding boxes, or quantitative measurements such as tumor diameters or fluid collection volumes. Its primary role is to supply precise, structured data that can be integrated into the radiology report, thereby enhancing its accuracy and clinical utility.

{\bf Diagnostic Classifier:} The diagnostic classifier agent synthesizes imaging-derived features into diagnostic assessments and preliminary recommendations. Functioning as an AI-driven “second opinion,” it provides interpretive insights that complement radiologist expertise. Its performance is evaluated using standard classification metrics as well as concordance with the final impression documented in radiologists’ reports.

{\bf Clinical Report Composer:} 
The clinical report composer serves as the central LLM agent responsible for composing the radiology report in natural language. It integrates inputs from across the pipeline, include structured findings from feature extractors, quantitative measurements from the segmentation agent, diagnostic insights, patient metadata, and other contextual information. Guided through structured prompting, this agent produces a coherent and clinical formatted report, typically organized into standardized sections such as "findings" and "impression".

{\bf Quality Assurance Agent (\emph{with} Human-in-the-Loop):} To ensure reliability and trustworthiness, the framework incorporates a quality assurance agent as a dedicated quality-control stage. This agent re-examines the generated report, cross-validating its contents against the source images and intermediate outputs produced by other agents. Its primary role is to detect inconsistencies, such as unsupported findings or omissions of clinically relivant information. The human-in-the-loop design ensures that the quality assurance agent can either consult an expert radiologist during the review process or incorporate feedback from radiologist annoutations, thereby aligning automated reporting with expert clinical judgment.

{\bf Evaluation Agent (Judge):} In addition to the internal quality assurance review, the framework incorporates an evaluation agent designed specifically for benchmarking purposes. Unlike agents involved in report generation, this component operates independently, assessing the final report and assigning scores across multiple quality dimensions. Beyond benchmarking, the evaluation agent can also serve as a reward model within reinforcement learning setups, guiding system optimization. Validation of this agent is performed by correlating its scoring outputs with expert radiologist evaluations on a subset of reports, leveraging recent evidence that LLM-based evaluators can achieve inter-rater reliability comparable to human reviewers. Within the benchmark, the evaluation agent thus provides an automated, scalable, and robust mechanism for comparing different models and configurations.

{\bf Orchestrator:} At the core of the framework lies the orchestrator agent, responsible for managing the sequence of operations, coordinating information flow, and ensuring overall coherence of the pipeline. It determines which specialized agents to invoke and in what order, adapting dynamically to the requirements of each case. In addition to coordination, the orchestrator performs critical validation checks to verify that outputs from individual agents are logically consistent, clinically plausible, and aligned with the intended reporting structure.

\section{Evaluation Protocols and Results}

We evaluate the proposed framework at both the agent and overall global levels, combining LLM-based evaluation methods with conventional classification and segmentation metrics. At the agent level, organ detection and modality classification are measured by accuracy and confusion matrices; content extraction by precision, recall, and F1-score. Segmentaion by Dice, IoU, and sensitivity/precision. Metadata processing by accuracy against ground truth text via LLM-as-judge. Diagnostic classification by accuracy, ROC AUC, free-text parsing, and concordance with radiologist impression. Report generation by ROUGE, clinical accuracy (CheXpert-based accuracy \cite{babar2021evaluating}), readability/coherence (LLM-based or Flesch-Kincaid Reading Grade \cite{picton2025assessing}), and radiologist-labeled clinically significant error rates. The reviewer by before/after improvement and planted-error precision/recall. Evaluation agent by calibration with expert scores (Spearman’s $\rho$)~\cite{dietz2025principles}. The orchestrator by pipeline success and efficiency. At the global level, performance is assessed through a composite Report Quality Score (emphasizing clinical accuracy over BLEU), a 0-100 benchmark score from the evaluation agent following VHELM~\cite{lee2024vhelm}, human preference rates from reader studies, clinician alignment metrics (factual correctness, uncertainty representation, and adherence to guidelines), and robustness in difficult or rephrased cases. We further simulate an RLHF loop, where pipeline adjustments are iteratively re-evaluated, yielding expected gains in factual accuracy and human preference. To ensure fairness, all systems are tested on the same standardized, held-out dataset with a shared scoring script for reproducible benchmarking.



\begin{figure}[t]
    \centering
   \includegraphics[width=0.98\linewidth]{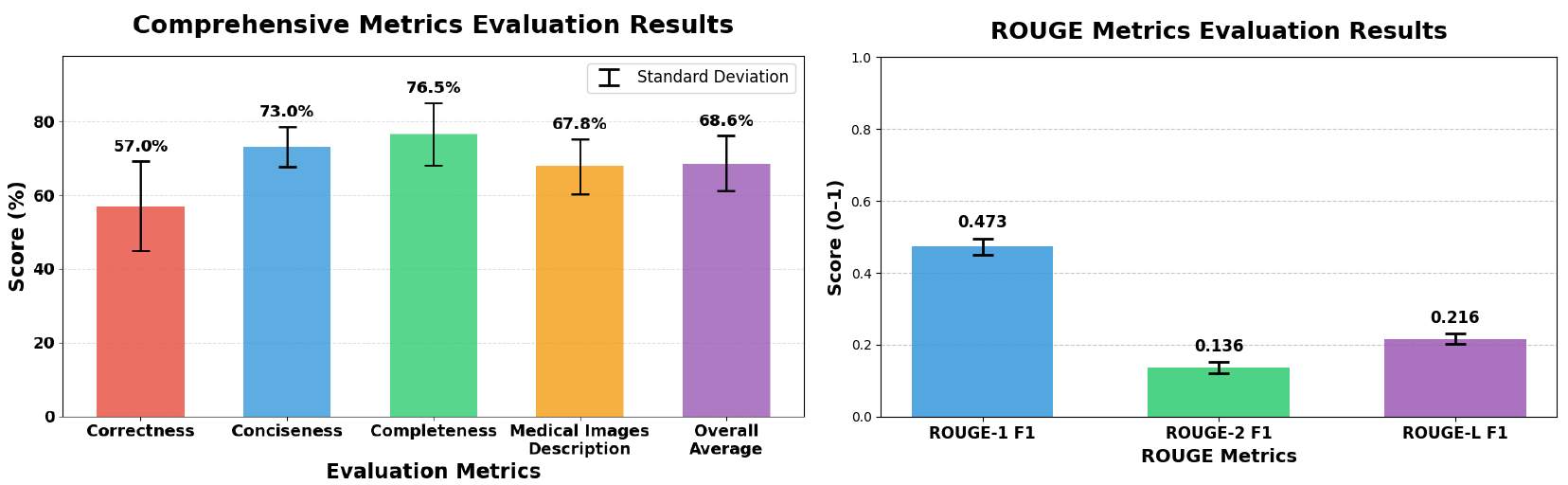}
    \caption{Performance results of the proposed agentic pipeline on the RHUH-GBM dataset.}
    \label{fig:agent_pipeline_results}
\end{figure}

In this study, we developed an adaptable pipeline designed for cross-dataset evaluation and applied it to the RHUH-GBM dataset~\cite{cepeda2023rio}, with an LLM serving as an automated judge. To establish reliable ground truth, reference radiology reports were created in collaboration with senior radiologists, ensuring clinically accurate annotations. The LLM evaluated system outputs along four dimensions: correctness (absence of medical errors), conciseness (clarity and brevity of expression), completeness (coverage of all key clinical findings), and image descriptions (quality of imaging-based descriptions)~\cite{van2024adapted}. As shown in Figure~\ref{fig:agent_pipeline_results}, the pipeline achieved an overall accuracy of 68.6\%. This also shows the limitation of traditional metrics like ROGUE in such complex tasks. Importantly, this evaluation was conducted without incorporating patient metadata such as tumor size or type, thereby testing the pipeline’s ability to independently infer these attributes. The results indicate that the system was largely successful in this regard, demonstrating robust performance in tumor detection and characterization, while also generating radiology reports that were comprehensive, clinically sound, and of high quality.

\section{Conclusion}
We presented Medical AI Consensus, a modular multi-agent framework that unifies radiology report generation and evaluation across LLMs and LVMs within a standardized, model-agnostic benchmark. Grounded in expert-authored reference reports and an LLM-as-judge, our evaluation spans correctness, conciseness, completeness, and image description. On the RHUH-GBM dataset, the system achieved an overall accuracy of 68.6\% without using patient metadata, highlighting the pipeline’s capacity to recover clinically salient information directly from images. The orchestrated, human-in-the-loop design promotes transparency, safety, and iterative refinement.

\bibliographystyle{plain}
\bibliography{ref}
\end{document}